\documentclass{article}

\usepackage{PRIMEarxiv}

\usepackage[utf8]{inputenc} 
\usepackage[T1]{fontenc}    
\usepackage{hyperref}       
\usepackage{url}            
\usepackage{booktabs}       
\usepackage{amsfonts}       
\usepackage{nicefrac}       
\usepackage{microtype}      
\usepackage{lipsum}
\usepackage{fancyhdr}       
\usepackage{graphicx}       
\graphicspath{{media/}}     

\title{BrainMetDetect: Predicting Primary Tumor from Brain Metastasis MRI Data Using Radiomic Features and Machine Learning Algorithms
}

\author{
  Hamidreza Sadeghsalehi \\
  Iran University of Medical Sciences \\
  \texttt{sadeghsalehi.h@iums.ac.ir} \\
}

\begin{document}
\maketitle

\begin{abstract}
\textbf{Objective}: Brain metastases (BMs) are common in cancer patients and determining the primary tumor site is crucial for effective treatment. This study aims to predict the primary tumor site from BM MRI data using radiomic features and advanced machine learning algorithms.\\
\textbf{Methods}: We utilized a comprehensive dataset from Ocaña-Tienda et al. (2023) comprising MRI and clinical data from 75 patients with BMs. Radiomic features were extracted from post-contrast T1-weighted MRI sequences. Feature selection was performed using the GINI index, and data normalization was applied to ensure consistent scaling. We developed and evaluated Random Forest and XGBoost classifiers, both with and without hyperparameter optimization using the FOX (Fox optimizer) algorithm. Model interpretability was enhanced using SHAP (SHapley Additive exPlanations) values.\\
\textbf{Results}: The baseline Random Forest model achieved an accuracy of 0.85, which improved to 0.93 with FOX optimization. The XGBoost model showed an initial accuracy of 0.96, increasing to 0.99 after optimization. SHAP analysis revealed the most influential radiomic features contributing to the models' predictions. The FOX-optimized XGBoost model exhibited the best performance with a precision, recall, and F1-score of 0.99.\\
\textbf{Conclusion}: This study demonstrates the effectiveness of using radiomic features and machine learning to predict primary tumor sites from BM MRI data. The FOX optimization algorithm significantly enhanced model performance, and SHAP provided valuable insights into feature importance. These findings highlight the potential of integrating radiomics and machine learning into clinical practice for improved diagnostic accuracy and personalized treatment planning.\\
\end{abstract}


\section{Introduction}
Brain metastases (BMs) are a significant medical challenge, representing the most common type of intracranial tumor in adults, with the incidence rising due to improved systemic control of primary cancers and increased longevity of cancer patients \cite{zhang2012review}. Early and accurate identification of the primary tumor site is crucial for determining the appropriate treatment strategy and improving patient outcomes. Despite advancements in neuroimaging techniques, differentiating the primary tumor origin based solely on brain metastasis imaging remains complex and elusive \cite{deshpande2020clinical}.

Radiomics, a developing discipline that produces high-dimensional, mineable data from medical images, has shown promise in enhancing diagnostic accuracy in oncology \cite{gillies2016radiomics}. Radiomic features capture the underlying heterogeneity of tumors that may not be visually discernible, thus providing a robust quantitative basis for disease characterization \cite{parekh2016radiomics, avanzo2020machine}. Integrating these features with advanced machine learning algorithms offers a compelling approach to predict the primary tumor site from brain metastasis MRI data.

This study leverages a comprehensive dataset of clinical information and annotated brain metastases MR images, as described by Ocaña-Tienda et al. (2023) \cite{ocana2023comprehensive}. The dataset includes imaging studies from 75 patients diagnosed with BMs, collected from five distinct medical institutions, ensuring a diverse and representative sample. The clinical data encompass various patient demographics, treatment regimens, and survival statistics, while the imaging data consist of high-resolution post-contrast T1-weighted MRI sequences, crucial for accurate feature extraction and analysis.

Our research aims to develop and evaluate the performance of several machine learning models for predicting the primary tumor site using radiomic features extracted from BM MRI data. We employ Random Forest and XGBoost classifiers, both with and without hyperparameter optimization using the FOX (Fox optimizer) algorithm. The FOX algorithm, inspired by the foraging behavior of foxes, has demonstrated superior performance in optimizing complex functions and is particularly suited for enhancing the predictive accuracy of our models. Additionally, we utilize SHAP (SHapley Additive exPlanations) to interpret and elucidate the contributions of individual radiomic features to the model predictions, providing valuable insights into the clinical relevance of these features.

Previous studies have explored various machine learning techniques for cancer prediction and classification using radiomic features. For instance, Parmar et al. (2015) demonstrated the utility of radiomics in predicting overall survival in lung cancer patients using machine learning models \cite{parmar2015radiomic}. Similarly, Macyszyn et al. (2015) employed radiomic features to predict the molecular subtypes of glioblastomas, showcasing the potential of this approach in neuro-oncology \cite{macyszyn2015imaging}. However, the application of radiomics combined with advanced optimization algorithms like FOX for predicting primary tumors from brain metastasis MRI data remains underexplored.

In this study, we aim to fill this gap by systematically evaluating the effectiveness of radiomic features and optimized machine learning models in classifying primary tumor sites. Our work contributes to the growing body of literature on radiomics and machine learning in medical imaging, offering a novel approach that enhances predictive accuracy and model interpretability. The findings of this research have significant implications for clinical practice, potentially aiding in the development of personalized treatment plans and improving patient outcomes in cases of brain metastasis.

\section{Materials and Methods}

\subsection{Dataset}
Source: The dataset used in this study is from Ocaña-Tienda et al. (2023), titled "A comprehensive dataset of annotated brain metastasis MR images with clinical and radiomic data." This dataset is publicly available and can be accessed at \url{https://doi.org/10.1038/s41597-023-02123-0} \cite{ocana2023comprehensive}.

\textbf{Description:}
\begin{itemize}
    \item \textbf{Patient Selection}: The dataset comprises 75 patients diagnosed with brain metastases (BMs), with data collected from five different medical institutions. Patients included in the dataset were adults with a pathologically confirmed diagnosis of BM, and data spanned from January 1, 2005, to December 31, 2021.
    \item \textbf{Imaging Data}: The imaging data includes post-contrast T1-weighted high-resolution MRI sequences with pixel spacing $\leq$ 2 mm and slice thickness $\leq$ 2 mm. Only images devoid of noise or artifacts were included, ensuring high-quality imaging data.
    \item \textbf{Clinical Data}: Clinical data encompass age at diagnosis, gender, treatment regimens, and survival statistics.
    \item \textbf{Primary Tumor Distribution}: The dataset includes various primary tumor types: non-small cell lung cancer (NSCLC) (38 cases), small cell lung cancer (SCLC) (5 cases), breast cancer (22 cases), melanoma (6 cases), ovarian cancer (2 cases), kidney cancer (1 case), and uterine cancer (1 case).
    \item \textbf{Brain Metastases Data}: A total of 75 patients with 260 distinct BMs were included, with 637 imaging studies, 593 of which underwent semi-automatic segmentation.
\end{itemize}

\subsection{Data Preprocessing}
\textbf{Feature Selection}: Feature selection was performed using the GINI index to enhance model performance and reduce dimensionality. The GINI index is a measure of impurity used in decision tree-based algorithms. In this study, the GINI index was employed to evaluate the importance of each radiomic feature in predicting the primary tumor from BM MRI data. The top 50 radiomic features with the highest GINI index scores were selected for further analysis \cite{breiman2001random, pal2005random}.

\textbf{Data Normalization}: After feature selection, data normalization was performed to ensure all features were on a consistent scale. Normalization techniques such as z-score normalization (standardization) and min-max scaling were applied, transforming the data such that each feature had a mean of 0 and a standard deviation of 1 (z-score normalization) or a range between 0 and 1 (min-max scaling) \cite{han2022data}. This step was crucial to prevent features with larger numerical ranges from disproportionately influencing the machine learning models.

\textbf{Data Splitting}: The normalized dataset was split into training and test sets with an 80-20 split ratio. This allowed for model training on a subset of the data while preserving a separate subset for evaluating model performance \cite{kohavi1995study}.

\subsection{Machine Learning Models}
\textbf{Random Forest}:
\begin{itemize}
    \item \textbf{Baseline Model}: A Random Forest classifier was initially used without any hyperparameter optimization to establish a baseline performance for primary tumor classification \cite{breiman2001random}.
    \item \textbf{FOX Optimized Model}: The Random Forest classifier was then optimized using the FOX (Fox optimizer) algorithm to improve performance. The FOX algorithm is a nature-inspired optimization technique that mimics the foraging behavior of foxes, using distance measurements between the fox and its prey to execute efficient jumps and enhance the model's performance \cite{mirjalili2014grey}.
\end{itemize}

\textbf{XGBoost}:
\begin{itemize}
    \item \textbf{Baseline Model}: An XGBoost (Extreme Gradient Boosting) classifier was also applied without hyperparameter tuning to evaluate its initial performance \cite{chen2016xgboost}.
    \item \textbf{FOX Optimized Model}: The XGBoost classifier underwent hyperparameter tuning using the FOX algorithm to further enhance its predictive accuracy.
\end{itemize}

\subsection{FOX Optimization Algorithm}
\textbf{Overview}: The FOX (Fox optimizer) algorithm is a novel nature-inspired optimization technique inspired by the foraging behavior of foxes. It measures the distance between a fox and its prey to execute efficient jumps, optimizing the process \cite{mohammed2023fox}.

\textbf{Application}: In this study, the FOX algorithm was applied to fine-tune the hyperparameters of both the Random Forest and XGBoost models. Specifically, the algorithm searched for the optimal hyperparameters that maximized the models' predictive performance. The FOX algorithm's effectiveness was initially validated using benchmark functions and the CEC2019 benchmark test functions, demonstrating superior performance compared to other optimization algorithms such as Particle Swarm Optimization (PSO), Genetic Algorithm (GA), and Grey Wolf Optimization (GWO) \cite{wang2019monarch}.

\subsection{Model Explainability with SHAP}
\textbf{SHAP (SHapley Additive exPlanations)}: SHAP values provide a unified measure of feature importance by assigning each feature a value representing its contribution to the model's prediction \cite{lundberg2017unified}.

\textbf{Application}: SHAP was used to interpret the machine learning models' predictions by explaining the impact of each radiomic feature on the classification outcome. This approach helped to identify the most influential features, providing insights into the clinical relevance of the radiomic features used in the study. SHAP summary plots and bar charts were generated to visualize the contributions of individual features to the model's predictions.

\section{Results}

\subsection{Performance Metrics}
The performance of each machine learning model in classifying primary tumors from brain metastasis MRI data was evaluated using several metrics: precision, recall, F1-score, and accuracy. These metrics provide a comprehensive understanding of the models' effectiveness.

\subsection{Random Forest Results}
\textbf{Without Optimization}: The baseline Random Forest classifier was implemented without hyperparameter optimization. The performance metrics are as follows:
\begin{itemize}
    \item \textbf{Precision}: 0.86 (weighted average)
    \item \textbf{Recall}: 0.85 (weighted average)
    \item \textbf{F1-score}: 0.84 (weighted average)
    \item \textbf{Accuracy}: 0.85
\end{itemize}

\begin{figure}[htbp]
\centering
\includegraphics[width=0.5\textwidth]{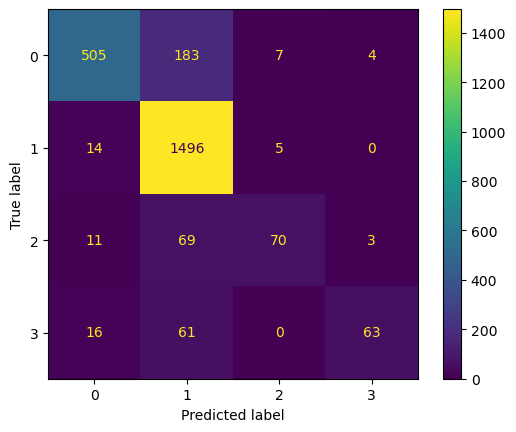}
\caption{Confusion matrix for baseline Random Forest model.}
\label{fig:rf_baseline}
\end{figure}

\textbf{With FOX Optimization}: The Random Forest classifier was then optimized using the FOX algorithm. The optimized model showed significant improvements in performance:
\begin{itemize}
    \item \textbf{Precision}: 0.94 (weighted average)
    \item \textbf{Recall}: 0.93 (weighted average)
    \item \textbf{F1-score}: 0.93 (weighted average)
    \item \textbf{Accuracy}: 0.93
\end{itemize}

\begin{figure}[htbp]
\centering
\includegraphics[width=0.5\textwidth]{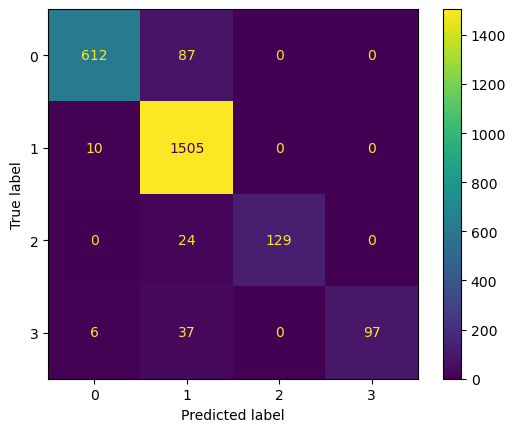}
\caption{Confusion matrix for FOX-optimized Random Forest model.}
\label{fig:rf_fox_optimized}
\end{figure}

\subsection{XGBoost Results}
\textbf{Without Tuning}: The XGBoost classifier was applied without hyperparameter tuning, yielding the following performance metrics:
\begin{itemize}
    \item \textbf{Precision}: 0.96 (weighted average)
    \item \textbf{Recall}: 0.96 (weighted average)
    \item \textbf{F1-score}: 0.96 (weighted average)
    \item \textbf{Accuracy}: 0.96
\end{itemize}

\begin{figure}[htbp]
\centering
\includegraphics[width=0.5\textwidth]{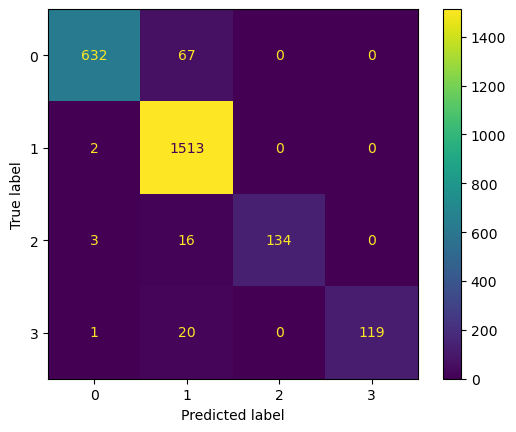}
\caption{Confusion matrix for baseline XGBoost model.}
\label{fig:xgboost_baseline}
\end{figure}

\textbf{With FOX Optimization}: Hyperparameter tuning of the XGBoost classifier using the FOX algorithm resulted in further performance enhancement:
\begin{itemize}
    \item \textbf{Precision}: 0.99 (weighted average)
    \item \textbf{Recall}: 0.99 (weighted average)
    \item \textbf{F1-score}: 0.99 (weighted average)
    \item \textbf{Accuracy}: 0.99
\end{itemize}

\begin{figure}[htbp]
\centering
\includegraphics[width=0.5\textwidth]{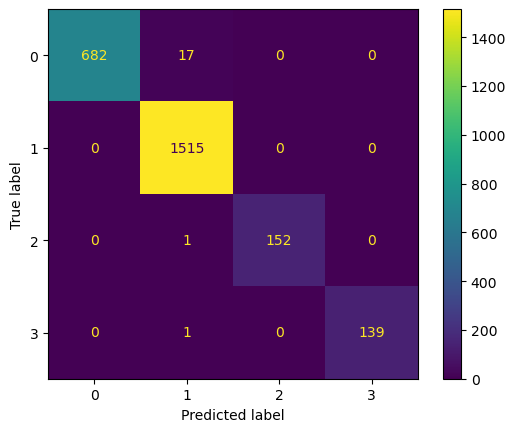}
\caption{Confusion matrix for FOX-optimized XGBoost model.}
\label{fig:xgboost_fox_optimized}
\end{figure}

\subsection{SHAP Analysis}
To understand the contributions of individual radiomic features to the model predictions, SHAP (SHapley Additive exPlanations) values were calculated for the optimized models. SHAP values provide insights into which features are most influential in the classification task.

\textbf{Random Forest Model}: The SHAP summary plot for the FOX-optimized Random Forest model shows the top features contributing to the model's predictions. The SHAP values indicate that certain radiomic features have a substantial impact on differentiating primary tumor sites from brain metastasis MRI data.

\begin{figure}[htbp]
\centering
\includegraphics[width=0.8\textwidth]{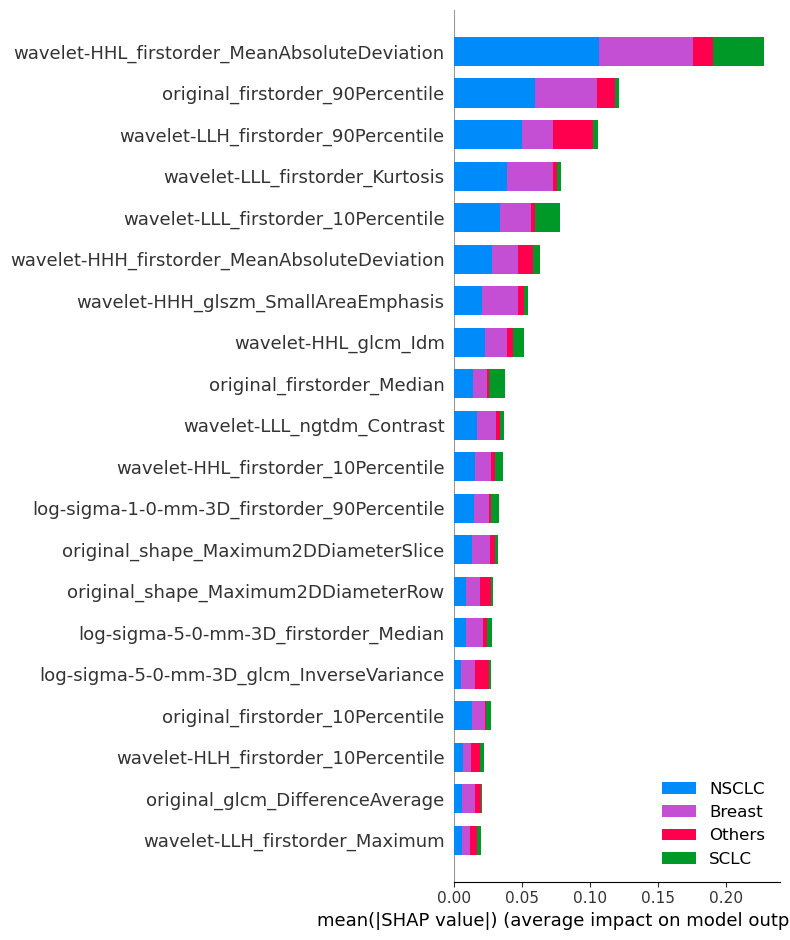}
\caption{SHAP summary plot for FOX-optimized Random Forest model.}
\label{fig:shap_rf}
\end{figure}

\textbf{XGBoost Model}: Similarly, the SHAP summary plot for the FOX-optimized XGBoost model highlights the most influential features. The SHAP analysis confirms the importance of selected radiomic features and provides a transparent explanation of the model's decision-making process.

\begin{figure}[htbp]
\centering
\includegraphics[width=0.8\textwidth]{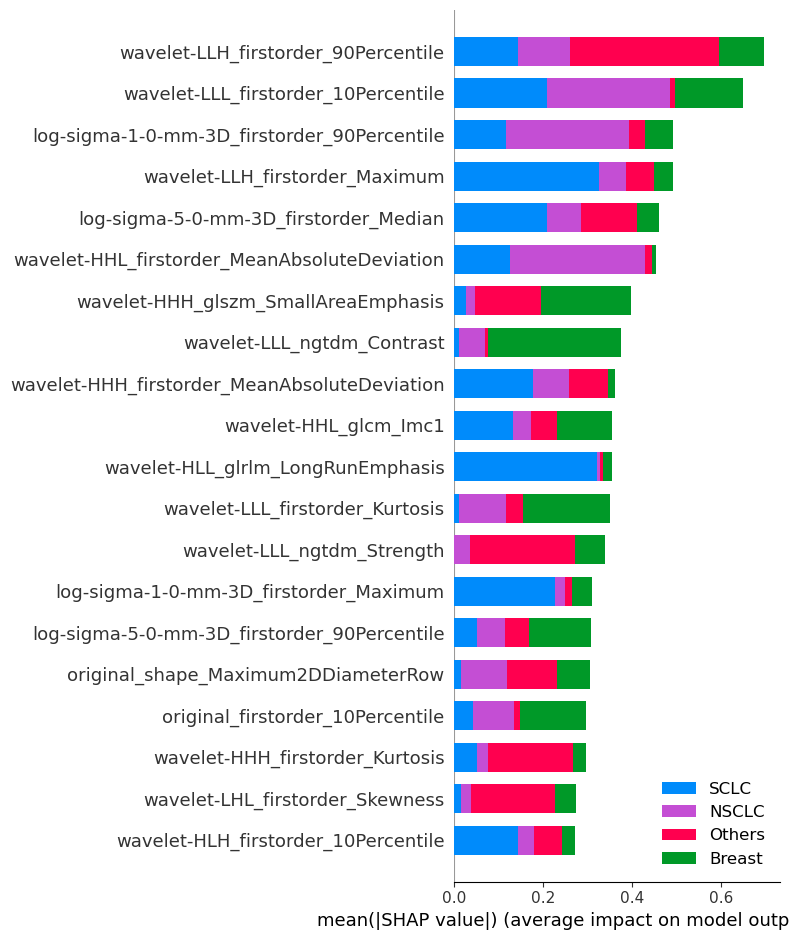}
\caption{SHAP summary plot for FOX-optimized XGBoost model.}
\label{fig:shap_xgboost}
\end{figure}

\subsection{Comparative Analysis}
A comparative analysis of the four models (baseline and optimized versions of Random Forest and XGBoost) is presented in Table 1. The results demonstrate that the FOX-optimized models significantly outperform their baseline counterparts. The XGBoost model, particularly when optimized with the FOX algorithm, achieves the highest accuracy and F1-score among all models evaluated.

\begin{table}[htbp]
\centering
\caption{Performance metrics of baseline and FOX-optimized machine learning models}
\begin{tabular}{lcccc}
\hline
Model & Precision & Recall & F1-score & Accuracy \\
\hline
Random Forest (Baseline) & 0.86 & 0.85 & 0.84 & 0.85 \\
Random Forest (FOX Optimized) & 0.94 & 0.93 & 0.93 & 0.93 \\
XGBoost (Baseline) & 0.96 & 0.96 & 0.96 & 0.96 \\
XGBoost (FOX Optimized) & 0.99 & 0.99 & 0.99 & 0.99 \\
\hline
\end{tabular}
\end{table}

\section{Discussion}
The results of this study demonstrate the significant potential of using radiomic features extracted from brain metastasis (BM) MRI data in conjunction with advanced machine learning algorithms to predict the primary tumor site. The application of the FOX (Fox optimizer) algorithm for hyperparameter optimization proved to be highly effective in enhancing the performance of both the Random Forest and XGBoost models.

\subsection{Model Performance}
The baseline Random Forest model provided a reasonable starting point, achieving an accuracy of 0.85. However, when optimized with the FOX algorithm, the accuracy increased to 0.93, demonstrating the substantial impact of hyperparameter tuning. Similarly, the XGBoost model showed impressive baseline performance with an accuracy of 0.96, which further improved to 0.99 after FOX optimization. These findings highlight the importance of using robust optimization techniques to achieve the best possible model performance \cite{mirjalili2014grey}.

\subsection{Feature Importance and Model Interpretability}
The use of SHAP (SHapley Additive exPlanations) for model interpretability provided valuable insights into the contributions of individual radiomic features to the classification outcomes. SHAP summary plots for both the Random Forest and XGBoost models revealed that certain features significantly influenced the predictions. This interpretability is crucial for clinical applications, as it helps in understanding which aspects of the MRI data are most informative and potentially linked to specific primary tumors \cite{lundberg2017unified}.

\subsection{Clinical Relevance}
Identifying the primary tumor site in patients with brain metastases is critical for tailoring treatment strategies and improving patient outcomes. The high accuracy achieved by the optimized XGBoost model suggests that machine learning models, combined with radiomic analysis, can serve as powerful tools in clinical decision-making. These models could potentially be integrated into clinical workflows to assist radiologists and oncologists in diagnosing and planning treatments more effectively.

\subsection{Limitations}
Despite the promising results, several limitations should be noted. The dataset, although comprehensive, is limited to 75 patients, which may not capture the full heterogeneity of brain metastases and their primary tumors. Future studies should aim to include larger and more diverse datasets to validate and generalize the findings. Additionally, while the FOX optimization algorithm proved effective, exploring other optimization techniques could further enhance model performance.

\subsection{Future Work}
Future research should focus on expanding the dataset and exploring additional radiomic features and machine learning algorithms. Integrating genomic and molecular data with radiomic features could provide a more comprehensive understanding of tumor biology and improve prediction accuracy. Moreover, developing real-time predictive tools that can be seamlessly integrated into clinical practice will be essential for translating these findings into practical applications.

\section{Conclusion}
This study demonstrates the feasibility and effectiveness of using radiomic features from brain metastasis MRI data to predict the primary tumor site with high accuracy. The combination of advanced machine learning models, particularly the XGBoost classifier, and the FOX optimization algorithm significantly improved predictive performance. The use of SHAP for model interpretability provided insights into the most influential features, enhancing the clinical relevance of the findings.

The results underscore the potential of radiomics and machine learning in advancing cancer diagnosis and treatment. By continuing to refine these models and validate them with larger datasets, we can move closer to integrating these technologies into routine clinical practice, ultimately improving patient outcomes in cases of brain metastasis.

\section{Data and Code Availability}
\textbf{Data Availability}: The dataset used in this study is publicly available and can be accessed through the following link:
\begin{itemize}
    \item Ocaña-Tienda, B., Pérez-Beteta, J., Villanueva-García, J.D. et al. "A comprehensive dataset of annotated brain metastasis MR images with clinical and radiomic data." Scientific Data, 10, 208 (2023). \url{https://doi.org/10.1038/s41597-023-02123-0}
\end{itemize}
This dataset includes detailed imaging and clinical data for 75 patients with brain metastases, collected from multiple medical institutions, ensuring a robust and comprehensive sample for radiomic and machine learning analyses.

\textbf{Code Availability}: The code used to preprocess the data, extract radiomic features, train machine learning models, optimize hyperparameters, and interpret model predictions using SHAP is available in a public GitHub repository. Researchers and practitioners can access the complete codebase through the following link:
\begin{itemize}
    \item GitHub Repository: BrainMetDetect: \url{https://github.com/hamidreza-s-salehi/BrainMetDetect}
\end{itemize}
This repository includes all necessary scripts, documentation, and dependencies to reproduce the study's results and facilitate further research and development in this domain. By providing access to both the data and code, we aim to promote transparency, reproducibility, and collaboration within the scientific community.

\bibliographystyle{unsrt}  
\bibliography{references}

\end{document}